\documentclass[10pt,twocolumn,letterpaper]{article}

\usepackage{cvpr}              % To produce the CAMERA-READY version
\usepackage{graphicx}
\usepackage{amsmath}
\usepackage{amssymb}
\usepackage{amsthm}
\usepackage{booktabs}
\usepackage{comment}

\usepackage[pagebackref,breaklinks,colorlinks]{hyperref}

\usepackage[capitalize]{cleveref}
\crefname{section}{Sec.}{Secs.}
\Crefname{section}{Section}{Sections}
\Crefname{table}{Table}{Tables}
\crefname{table}{Tab.}{Tabs.}

\theoremstyle{definition}

\def\confName{NeuroVision - CVPR}
\def\confYear{2022}

\begin{document}

\title{Hyper-Universal Policy Approximation:\\Learning to Generate Actions  from  a Single Image using Hypernets}

\author{Dimitrios C. Gklezakos, Rishi Jha, Rajesh P. N. Rao \thanks{
 This work is supported by the Defense Advanced Research Projects Agency (DARPA) Contract No. HR001120C0021, 
 National Science Foundation (NSF) Grant \#EEC-1028725, a Weill Neurohub Investigator grant and a grant from the Templeton World Charity Foundation.}\\
Paul G. Allen Center for Computer Science \& Engineering\\
University of Washington, Seattle, WA 98195\\
{\tt\small \{gklezd, rjha01, rao\}@cs.washington.edu}
}

\maketitle

%%%%%%%%% ABSTRACT
\begin{abstract}
Inspired by Gibson's notion of object affordances in human vision, we ask the question: how can an agent learn to predict an entire action policy for a novel object or environment given only a single glimpse? To tackle this problem, we introduce the concept of Universal Policy Functions (UPFs) which are state-to-action mappings that generalize not only to new goals but most importantly to novel, unseen environments. Specifically, we consider the problem of efficiently learning such policies for agents with limited computational and communication capacity, constraints that are frequently encountered in edge devices. We propose the Hyper-Universal Policy Approximator (HUPA), a hypernetwork-based model to generate small task- and environment-conditional policy networks from a single image, with good generalization properties. Our results show that HUPAs significantly outperform an embedding-based alternative for generated policies that are size-constrained. Although this work is restricted to a simple map-based navigation task, future work includes applying the principles behind HUPAs to learning more general affordances for objects and environments.
\end{abstract}

%%%%%%%%% BODY TEXT
\section{Introduction}
The American psychologist James Gibson first proposed the idea of {\em object affordances} \cite{Gibson1966}, namely, that an object is perceived not only by the object's visual features, but also by the potential motor actions it affords. The idea that a single image of an object can suggest action plans to achieve particular goals has important implications for computer vision-based robotic agents: it can allow zero-shot generalization of learned skills to new objects and environments.  Can such a capability be realized by computer vision systems?

In reinforcement learning, action plans are formalized in terms of a policy function $\pi(s)\rightarrow a$ that maps the current state $s$ of the agent to the desired action $a$. Policy-based techniques, widely used in reinforcement learning \cite{NIPS1999_464d828b}, aim to find strategies that maximize the expected reward received from the environment. 
In most realistic scenarios, agents are required to be adept in many different tasks and good policies often exploit shared structure defined by task similarity and the dynamics of the environment. For example, in the case of navigation, policies that guide an agent to two different goals in close proximity should agree for most states $s$. Similarly, policies for different but closely-related environments should also be related. To extrapolate policies in novel environments, humans often rely on visual cues. In navigation tasks, humans can utilize a visual map to navigate in a previously unknown space. 

To formalize this correspondence between visual (and potentially other) auxiliary information and action policies, we propose the concept of a universal policy function (UPF) $\pi_{E}(s,g)$, where $g$ is the goal and $E$ a description of the environment, obtained, for example, from a visual map. UPFs are related to general value functions  \cite{conf/atal/SuttonMDDPWP11,pmlr-v37-schaul15},  but besides generalizing value functions to novel goals, we additionally tackle the challenge of transferring policies to novel environments, a new dimension which introduces significant complexity.

Although UPFs can be approximated by any generic neural network, training such models based on a limited set of goal and environment samples can be challenging. An agent with limited computational capacity and communication bandwidth faces another challenge; storing or transmitting a monolithic model with lots of parameters can be unwieldy or even impossible. Such constraints are frequently encountered in designing edge devices.

In this paper, we propose Hyper-Universal Policy Approximators (HUPAs) which, to our knowledge, is the first framework for mapping visual information, e.g., a single map image, to universal policy functions for agents with computational and communication constraints. HUPAs leverage the modularity properties of hypernetworks \cite{NEURIPS2020_75c58d36} to generate small environment-conditional, policy functions from an image (or other auxiliary information). These policy functions can then be easily transmitted to the agent and fine-tuned. We compare HUPAs to the traditional embedding alternative and demonstrate a significant performance gap between the two approaches when the generated policy network is size-constrained. While we focus on a simple map-based navigation task, our results are the first to highlight the advantage of hypernetworks in representing UPFs.

\section{Universal Policy Approximation}
In the context of a Markov Decision Process (MDP), we denote a UPF by $\pi_{E}(s,g)\rightarrow a$, where $g$ is the goal state, $s$ the current state, $a$ the chosen action, and $E$ a context vector that serves as a description of the environment. This description contains any auxiliary information that can be useful in conditioning the resulting policy. It can be given explicitly (e.g., image of a map for map-based navigation, an embedding-based natural language description, etc.) or implicitly (inferred from local observations such as images from a robot's on-board camera). The goal is to efficiently generate policies $\pi_E$ conditioned on a specific novel environment description $E$.

We make the following assumptions for our setting: \textbf{(a)} the agent/edge device is limited in computational capacity, \textbf{(b)} it operates within a specific context/environment $E$ and \textbf{(c)} has access to a device with large computational capacity on the cloud.

\begin{figure}
    \centering
    \begin{subfigure}[b]{0.25\textwidth}
    \includegraphics[width=\textwidth]{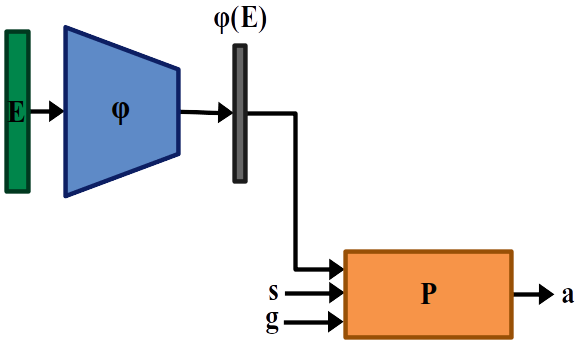}
    \caption{Embedding Approach}
    \label{fig:models:emb}
    \end{subfigure}
    \begin{subfigure}[b]{0.25\textwidth}
    \includegraphics[width=\textwidth]{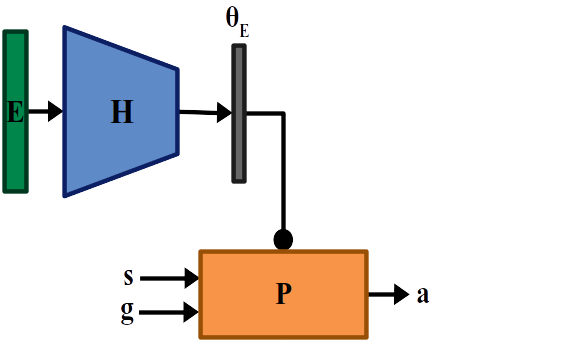}
    \caption{HUPA}
    \label{fig:models:hyp}
    \end{subfigure}
    \caption{\textbf{Embedding versus Hypernet-based Approaches for Predicting Action Policies from a Single Environmental Input.}}
    \label{fig:models}
\end{figure}

A straightforward way to approximate $\pi_{E}$ is to use a large network $\phi$ hosted on the cloud to obtain a context embedding $\phi(E)$. This embedding is then transmitted to the edge device. The edge device hosts a size-limited network $P$ that takes as input $x=[s,g,\phi(E)]$ and predicts the appropriate action for each $x$. We refer to this technique as the ``embedding'' approach. Note that instead of transmitting the whole embedding vector it suffices to transmit its projection to the first layer of $P$ which is usually smaller in size. A diagram for the embedding approach is shown in Figure \ref{fig:models:emb}.

As an alternative, we propose Hyper-Universal Policy Approximators (HUPAs), based on hypernetworks \cite{DBLP:journals/corr/HaDL16}. A hypernetwork is a network that generates parameters for another network called the primary network. The hypernetwork $H$, which is hosted on the cloud, takes as input the context vector and outputs a set of parameters $\theta_E = H(E)$. These parameters are then transmitted to the edge device and are used to parameterize a small primary network $P$. The HUPA model is shown in Figure \ref{fig:models:hyp}. Recent results on the modularity of hypernetworks show that they are provably more efficient approximators of functions of this form  compared to the embedding approach \cite{NEURIPS2020_75c58d36}.

\begin{figure}[h!]
    \centering
    \begin{subfigure}[b]{0.48\textwidth}
    \includegraphics[width=\textwidth]{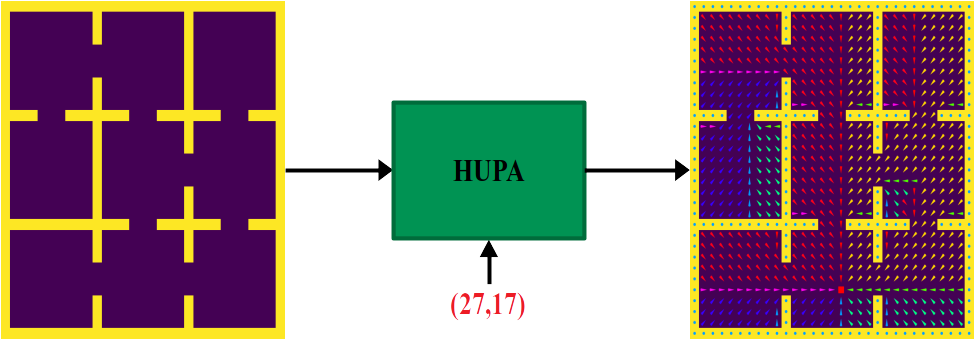}
    \caption{Known $E$ - Known $g$}
    \end{subfigure}
    
    \begin{subfigure}[b]{0.48\textwidth}
    \includegraphics[width=\textwidth]{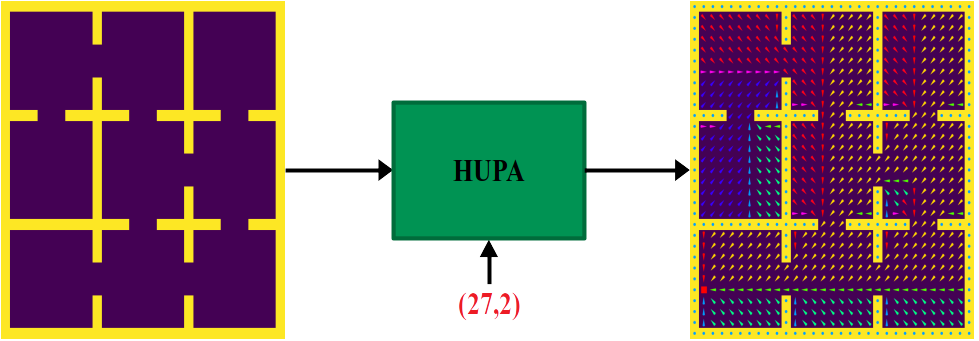}
    \caption{Known $E$ - Unknown $g$}
    \end{subfigure}
    
    \begin{subfigure}[b]{0.48\textwidth}
    \includegraphics[width=\textwidth]{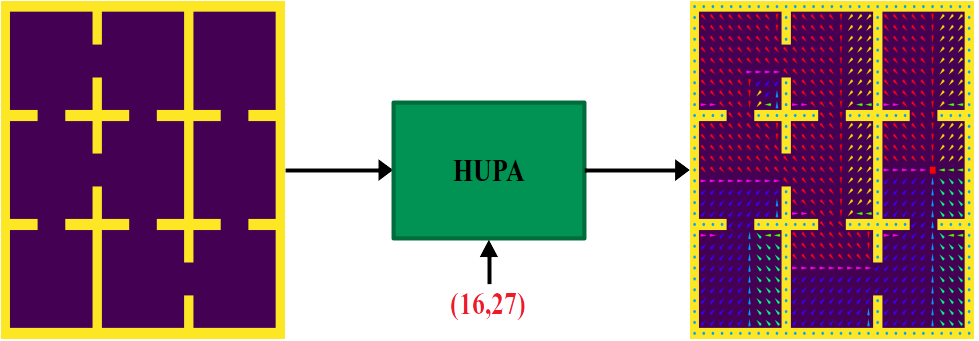}
    \caption{Unknown $E$ - Unknown $g$}
    \end{subfigure}
    
    \caption{\textbf{Zero-Shot Learning of Policies from Single Images:} After training on known $E$ and $g$ samples, a HUPA \textbf{(a)} predicts policies for novel goals \textbf{(b)} and novel environments \textbf{(c)} from a single input map image. The goal is marked by a red square on the right-side policy maps;  action vectors are color-coded by angle.}
    \label{fig:zeroshot}
\end{figure}

\section{Evaluation}
\subsection{Dataset and Metrics}
We evaluate HUPAs on a novel, artificially generated dataset based on a two-dimensional navigation task. The agent is given a map $E$, a starting coordinate $s$ and must navigate to a goal $g$. The maps consist of discretized tiles which are either open or occupied by a wall. The action space of the agent consists of $8$ actions/angles $a \in \{\frac{k\pi}{4}: k \in \{1,...,8\}\}$ that transfer the agent to one of the neighboring tiles (up, down, left, right and the diagonals) as long as they are open. Our goal is to learn a HUPA that, given previously unseen maps and goals, can generate policies that transfer the agent successfully to the target location. Since the action space is discrete, the generated primary network $P$ takes the form of a classifier. 

\subsubsection{Map Generation}
For our experiments we use a nine room generalization of the four-room environment commonly used in RL navigation scenarios. In this regime, each map $E$ is derived from a base map that contains rooms arranged in a $3\times 3$ grid separated by walls. Each room consists of a $9\times 9$ grid of open tiles, and neighbouring rooms are connected by doors in the center of the separating wall. There are twelve doors in total.
To generate a new map from the base map, three doors are chosen to be blocked off under the constraint that the space remains connected. This results in $164$ potential maps. From each map all possible start and goal coordinates are sampled to create the dataset. By representing states $s,g$ by their two-dimensional coordinates $(x,y)$ the maps can be conveniently represented as images, as we do in the following experiments. HUPAs function as zero-shot policy generators for previously unseen environments based on these single images (see Figure \ref{fig:zeroshot}).

\subsubsection{Reachability Ratio}
While canonical accuracy serves as a good measure of how close our generated policies are to the ground truth policy, it does not encapsulate the usability of a policy. The successful transition of the agent to the goal state relies on a sequence of several correctly predicted actions, a requirement that is not captured by plain accuracy. As an example a generated policy can leave a whole room disconnected from the goal, whereas most of the state-action pairs inside the room are predicted correctly (for examples of such generated policies, see Figure \ref{fig:graph_low_uu}).

A natural way to properly assess the quality of a generated policy is to evaluate the fraction of states from which it successfully guides the agent to the goal. We call this the Reachability Ratio (RR) metric.
To that end we construct an incidence graph where each state $s$ is represented by a node $n_s$. For each such state we add an edge $n_s\rightarrow n_{\pi_E(s,g)}$ that corresponds to the immediate transition that results from following the policy at $s$. Let $C_g$ be the connected component of this graph that contains $n_g$. Then $C_g$ contains all the nodes from which the policy successfully reaches the goal.
Hence the Reachability Ratio of a set $T$ of test goals is the average fraction of nodes in the target set ``accessible'' to each state in the map:
\begin{equation}
    \textrm{RR}(T) = \frac{1}{|S|}\sum_{g \in S}\frac{|T \cap C_g|}{|T|}
\end{equation}
where $C_g$ is computed via the reachability graph of $\pi_E$.

% \begin{comment}
% \begin{definition}[Reachability Ratio (RR)]
% Let $T \subset S$ be the subset of targets states in state set $S$, $E$ be a map, and $G$ be the reachability graph such that $C_g \subset G$ is the connected component containing state $g$ within $G$.
% Then, the Reachability Ratio of set $T$ on map $E$ and policy $\pi$ is:
% $$.$$
% \end{definition}
% \end{comment}

\subsection{Experiments}
For the embedding function $\phi$, we used a convolutional neural network with four residual blocks, followed by a fully connected bottleneck layer. An additional layer outputs the embedding vector $\phi(E)$. The hypernetwork had almost identical structure, except for an additional layer that generates the parameters $\theta_E$. For a fair comparison we adjust the size of the embedding vector accordingly, while keeping the size of the primary network fixed, to approximately equate the total number of parameters.

The primary network consists of three hidden layers with the same number of neurons, followed by an output layer that predicts the movement direction. We compare the two approaches on primary networks with $16,32,64,128$ and $256$ neurons. The resulting models have $315K ,544K ,1.4M ,4.7M$ and $17.6M$ parameters respectively. We used $50$ maps and $40\%$ of the states at random as our training set. We used early stopping on a validation set of $5$ novel maps and $10\%$ novel goals, with patience equal to $10$. For evaluation we used $20$ test maps and the remaining goals.

\begin{figure}[h!]
    \centering
    \begin{subfigure}[b]{0.45\textwidth}
    
    \centering
    \begin{subfigure}[b]{0.4\textwidth}
    \centering
    \includegraphics[width=\textwidth]{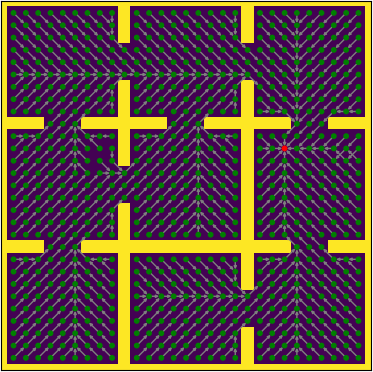}
    \caption{Known $E$ - Known $g$}
    \label{fig:graph_kk}
    \end{subfigure}
    \begin{subfigure}[b]{0.4\textwidth}
    \centering
    \includegraphics[width=\textwidth]{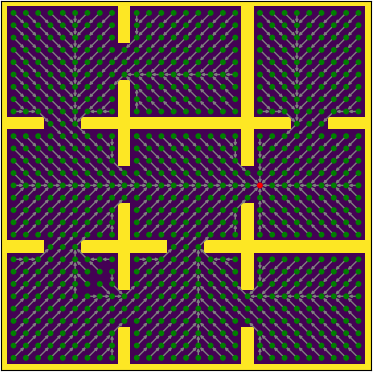}
    \caption{Known $E$ - Unknown $g$}
    \label{fig:graph_ku}
    \end{subfigure}
    
    \end{subfigure}

    \begin{subfigure}[b]{0.45\textwidth}
    \centering
    \includegraphics[width=0.4\textwidth]{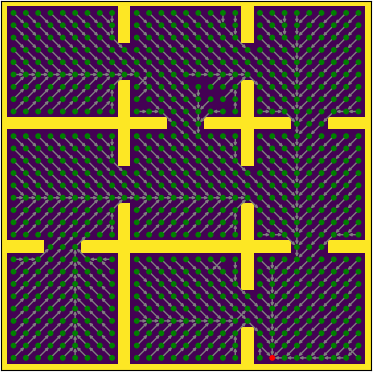}
    \includegraphics[width=0.4\textwidth]{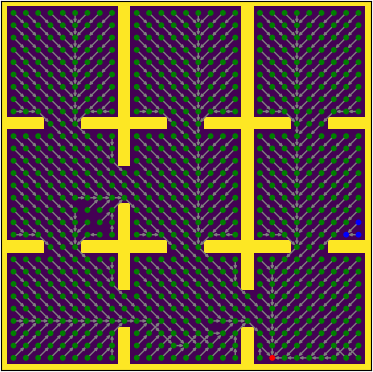}
    \caption{High quality policies: Unknown $E$ - Unknown $g$}
    \label{fig:graph_high_uu}
    \end{subfigure}
    
    \begin{subfigure}[b]{0.45\textwidth}
    \centering
    \includegraphics[width=0.4\textwidth]{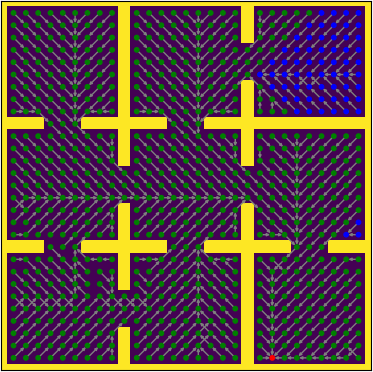}
    \includegraphics[width=0.4\textwidth]{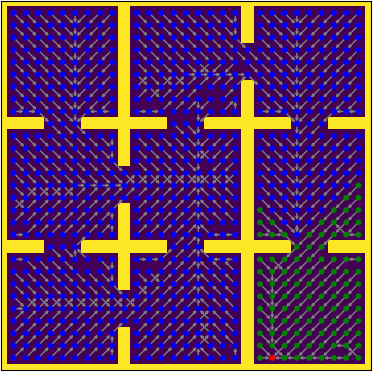}
    \caption{Low  quality policies: Unknown $E$ - Unknown $g$}
    \label{fig:graph_low_uu}
    \end{subfigure}
    
    \caption{\textbf{Reachability Graph Examples:} Examples of reachability graphs in policies predicted by the HUPA model for unseen maps and goals. Walls are denoted by yellow, open tiles by purple, while the red node marks the goal. Green nodes are states from which the goal is reachable under the predicted policy. Blue nodes do not reach the goal. The failure example in (d) (right) illustrates the importance of using reachability compared to accuracy, and also the potential for fixing the policy via few-shot RL.}
    \label{fig:reachability}
\end{figure}

Figure \ref{fig:reachability} shows examples of generated policies represented by their reachability graphs. Figures \ref{fig:graph_kk} and \ref{fig:graph_ku} show high-quality policies for known maps and known/unknown goals. Figure \ref{fig:graph_high_uu} shows successful zero-shot policy generation for unknown maps and goals. Finally Figure \ref{fig:graph_low_uu} shows examples of policies that fail to correctly understand the structure of the map, leaving whole rooms disconnected from the goal. However,  these rooms are disconnected due to only a few erroneous decisions. Fine-tuning the generated policies using reinforcement learning based on a few episodes could potentially correct these errors.

\begin{figure}[h!]
    \centering
    \includegraphics[width=0.5\textwidth]{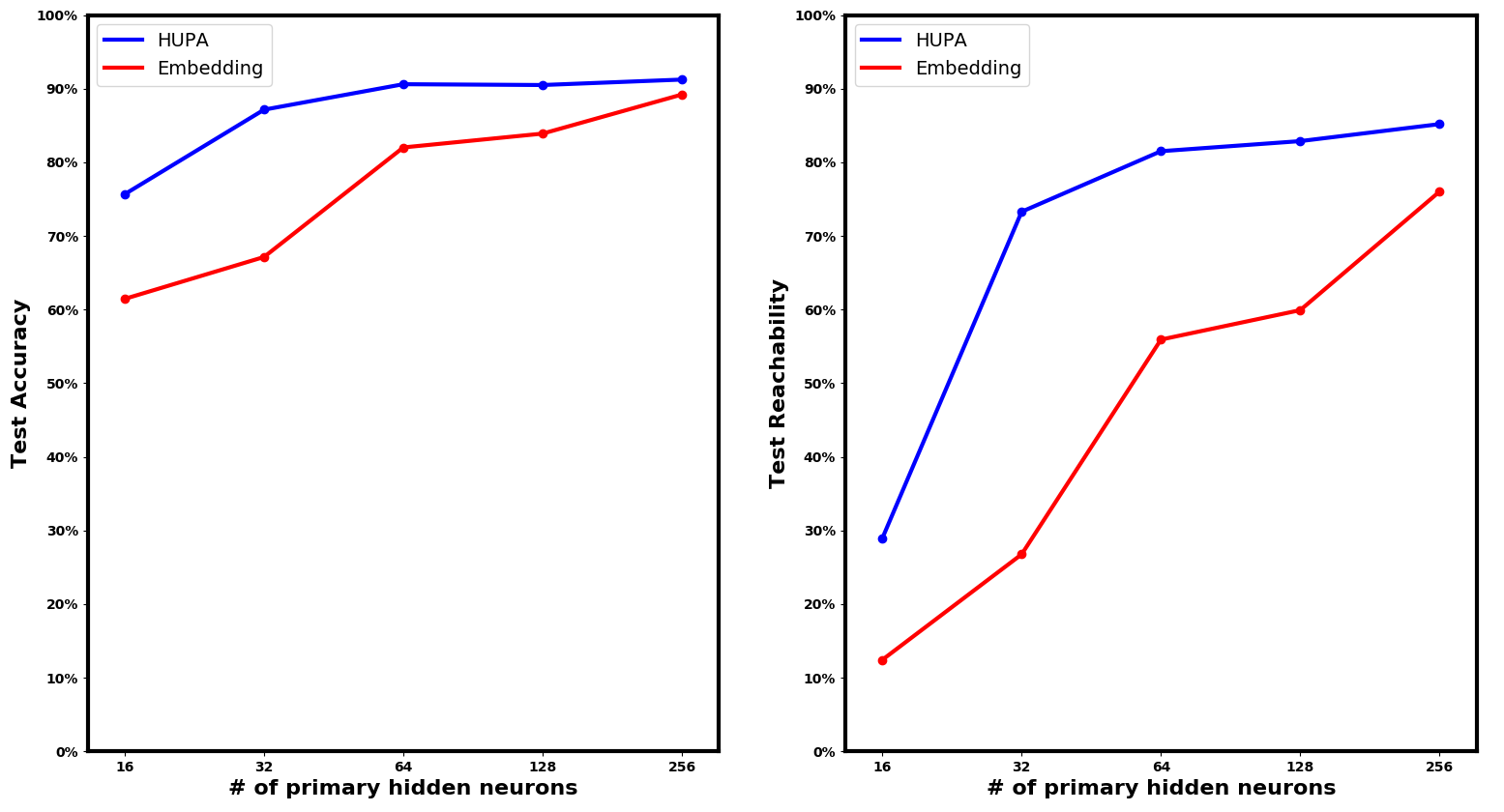}
    \caption{\textbf{Generalization to Novel Maps and Goals:} Performance of HUPAs on novel maps and goal compared to the embedding approach: policy accuracy \textbf{(left)}; goal reachability \textbf{(right)}.}
    \label{fig:main_result}
\end{figure}

We quantitatively compared HUPAs to the embedding alternative both in terms of accuracy and reachability. As seen in Figure \ref{fig:main_result}, HUPAs significantly outperform the baseline in both metrics. The effect is more pronounced for more constrained primary networks, with the embedding model approaching the performance of HUPAs as the parameter count becomes significantly higher.

\begin{figure}[h!]
    \centering
    \includegraphics[width=0.5\textwidth]{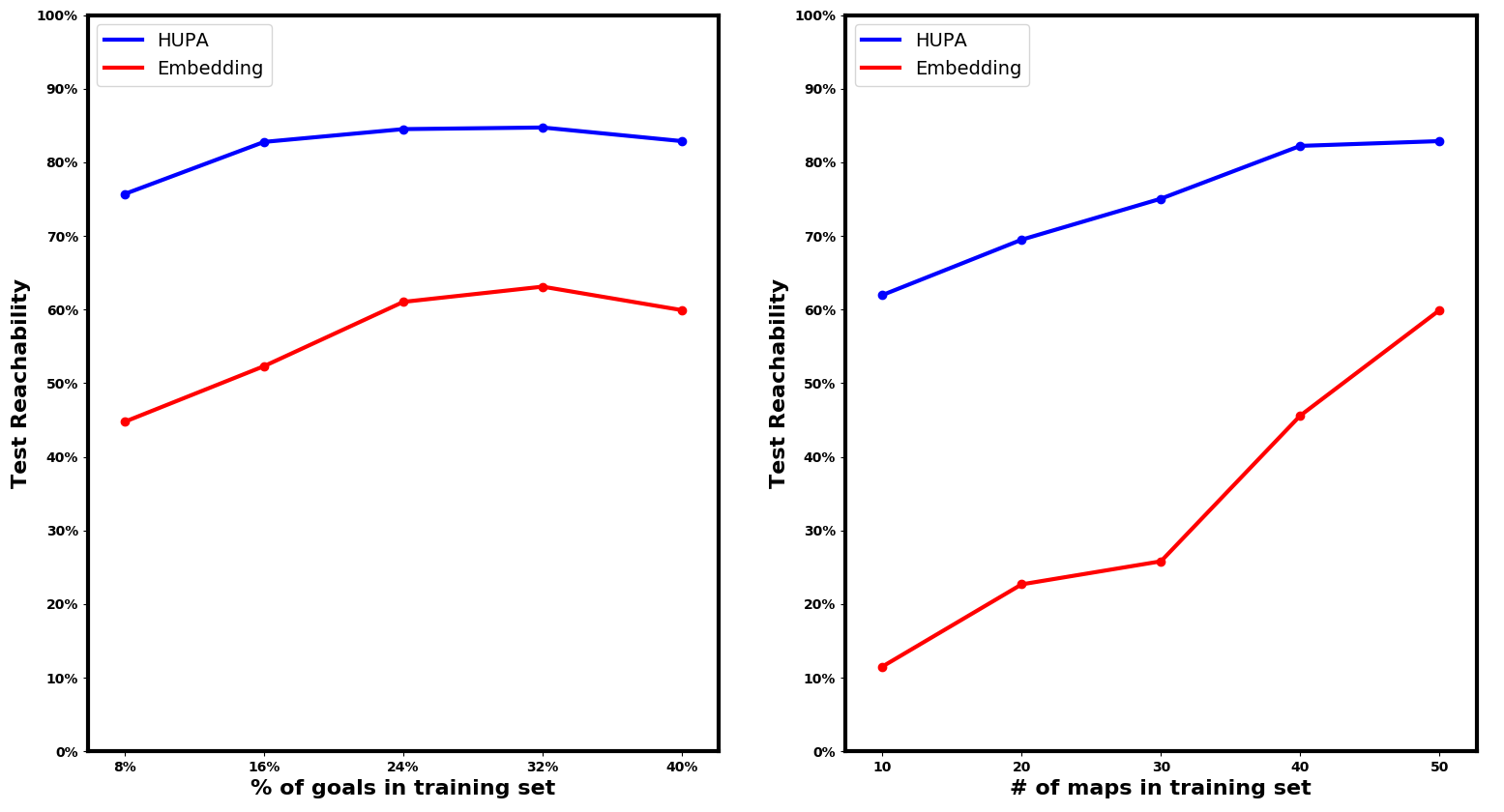}
    \caption{\textbf{Robustness to Map and Goal Sparsity in Training:} Goal reachability as a function of percentage of total possible goals \textbf{(left)} and maps \textbf{(right)} used for training.}
    \label{fig:density_result}
\end{figure}

To evaluate the robustness of our approach, we varied the number of training maps and goal states. We chose the $128$ neuron model for our primary architecture. As seen in Figure~\ref{fig:density_result}, the hypernetwork generalizes well even when the map/goal training set is sparse. HUPAs consistently outperform the embedding model in all sparsity settings.

\section{Conclusion}
We proposed HUPAs, a hypernetwork-based model for zero-shot generation of whole action policies from a single image, thereby suggesting a neural network implementation of Gibson's affordances. We demonstrated the significant advantage of HUPAs over the traditional embedding approach for size-constrained primary networks. Future work includes fine-tuning and meta-learning the generated policies, applying HUPAs to more complex settings and inferring environment descriptions directly from observations.

% Uncomment below for camera-ready version
%\vspace*{.05in}
%\noindent{\bf Acknowledgements}. Work supported by Templeton World Charity Foundation and DARPA (HR001120C0021).
%%%%%%%%% REFERENCES
{\small
\bibliographystyle{ieee_fullname}
\bibliography{hupa}
}

\end{document}